%% file: main.tex
\documentclass[nonacm,sigconf]{acmart}
\pdfoutput=1

\AtBeginDocument{%
  \providecommand\BibTeX{{%
    \normalfont B\kern-0.5em{\scshape i\kern-0.25em b}\kern-0.8em\TeX}}}

\usepackage{url}
\usepackage{makecell}
\usepackage{graphicx}  
\usepackage{multirow}
\usepackage{tabularx}
\usepackage{booktabs}
\usepackage{colortbl}
\usepackage{arydshln}
\usepackage{subcaption}
\usepackage{xcolor}
\usepackage{xspace}
\usepackage{hhline}
\usepackage[T1]{fontenc}
\usepackage[utf8]{inputenc}
\usepackage{babel}
\usepackage[font=small,labelfont=bf,tableposition=top]{caption}
\usepackage{booktabs}
\usepackage{threeparttable}
\definecolor{mygreen}{rgb}{0.29, 0.7, 0.48}

\begin{document}

\title{Animated Stickers: Bringing Stickers to Life with Video Diffusion}
\author{David Yan*, Winnie Zhang*, Luxin Zhang, Anmol Kalia, Dingkang Wang, Ankit Ramchandani, Miao Liu, Albert Pumarola, Edgar Sch{\"o}nfeld, Elliot Blanchard, Krishna Narni, Yaqiao Luo, Lawrence Chen, Guan Pang, Ali Thabet, Peter Vajda, Amy Bearman\textsuperscript{\textdagger}, Licheng Yu\textsuperscript{\textdagger}}
\affiliation{%
  \institution{GenAI, Meta}
  \city{Menlo Park, California}
  \country{USA}
}

\begin{abstract}
We introduce \textit{animated stickers}, a video diffusion model which generates an animation conditioned on a text prompt and static sticker image. 
Our model is built on top of the state-of-the-art Emu text-to-image model, with the addition of temporal layers to model motion. 
Due to the domain gap, i.e. differences in visual and motion style, a model which performed well on generating natural videos can no longer generate vivid videos when applied to stickers. 
To bridge this gap, we employ a two-stage finetuning pipeline -- first with weakly in-domain data, followed by human-in-the-loop (HITL) strategy which we term \textit{ensemble-of-teachers}.
It distills the best qualities of multiple teachers into a smaller student model. 
We show that this strategy allows us to specifically target improvements to motion quality while maintaining the style from the static image. 
With inference optimizations, our model is able to generate an eight-frame video with high-quality, interesting, and relevant motion in under one second.
\end{abstract}

\begin{teaserfigure}
  \centering
  \includegraphics[width=\linewidth]{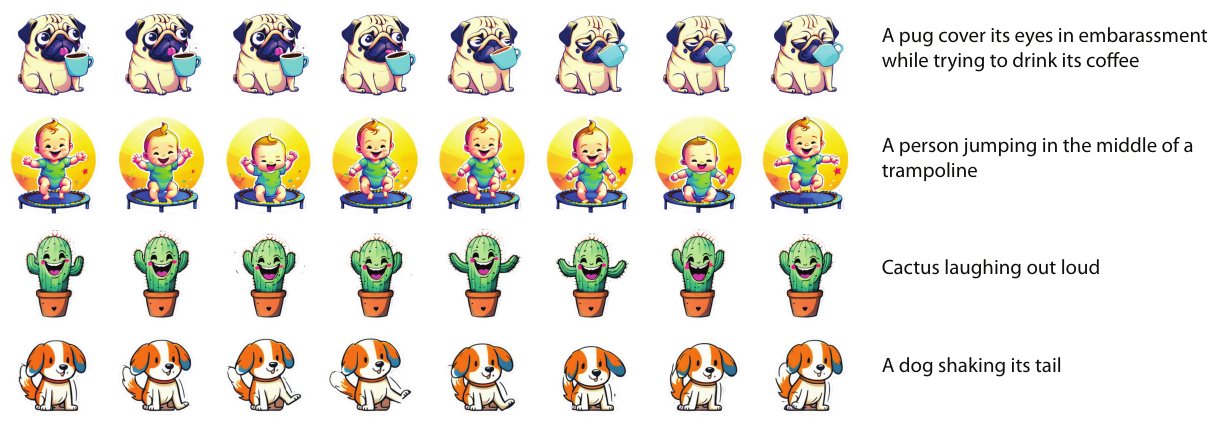}
  \caption{Examples of animated stickers generated by our model. Motions display a high degree of quality, consistency, expressiveness and relevance to the subject. Images are shown with transparent regions replaced by a white background.}
  \label{fig:teaser}
\end{teaserfigure}

\maketitle
\def\thefootnote{*}\footnotetext{Indicates equal contribution}
\def\thefootnote{\textdagger}\footnotetext{Corresponding authors}
\input{1.introduction}
\input{2.model_data}
\input{3.training_details}
\input{4.evaluation_results}

\section{Conclusion}
We presented our animated stickers model, which uses a spatiotemporal latent diffusion model conditioned on a text-image pair to animate sticker images. 
Our pretrain-to-production pipeline started with the Emu model, which was fine-tuned on a large set of natural videos, followed by in-domain datasets. 
We then use an ensemble-of-teachers HITL fine-tuning strategy to further improve the motion quality, consistency, and relevance. 
We use a number of architectural, distillation-based optimizations, and post-training optimizations to speed up the inference to one second per batch. 
We show that our fine-tuning strategy improves motion size and quality significantly over a model trained on natural videos only, demonstrating the effectiveness of the ensemble-of-teachers approach, and our other train-time improvements, such as middle-frame conditioning and motion bucketing. 
Our model is currently in public testing.

There are several areas for future work. 
First, our current model only outputs 8 frames, which limits the potential range of motion; increasing the number of frames while maintaining inference time would be an important improvement. 
Second, modifying model outputs to ensure smooth looping would improve user experience, since stickers are automatically looped for users, and large jumps between the first and last frame cause an unpleasant effect. 
Finally, there is still room to improve the overall quality of primary and secondary motion by expanding and further filtering datasets, tweaking model architecture, and further reducing quality loss in inference optimizations.

\begin{acks}
We would like to thank Anthony Chen, Ishan Misra, Mannat Singh, Rohit Girdhar, Andrew Brown, Saketh Rambhatla, Quentin Duval, Samaneh Azadi, Samyak Datta, Kapil Krishnakumar, Tsahi Glik, Jeremy Teboul, Shenghao Lin, Milan Zhou, Karthik Sivakumar, Ashley Ngo, Thai Quach, Jiabo Hu, Yinan Zhao, Bichen Wu, Ching-Yao Chuang, Arantxa Casanova Paga, Roshan Sumbaly, and Tali Zvi for their helpful discussions, guidance, and support which made this work possible.
\end{acks}

\bibliographystyle{ACM-Reference-Format}
\bibliography{bib}
\clearpage
\appendix
\onecolumn
\section{More animation examples} \label{more_animation_examples}
\begin{figure}[h]
    \centering
    \includegraphics[width=1.0\linewidth]{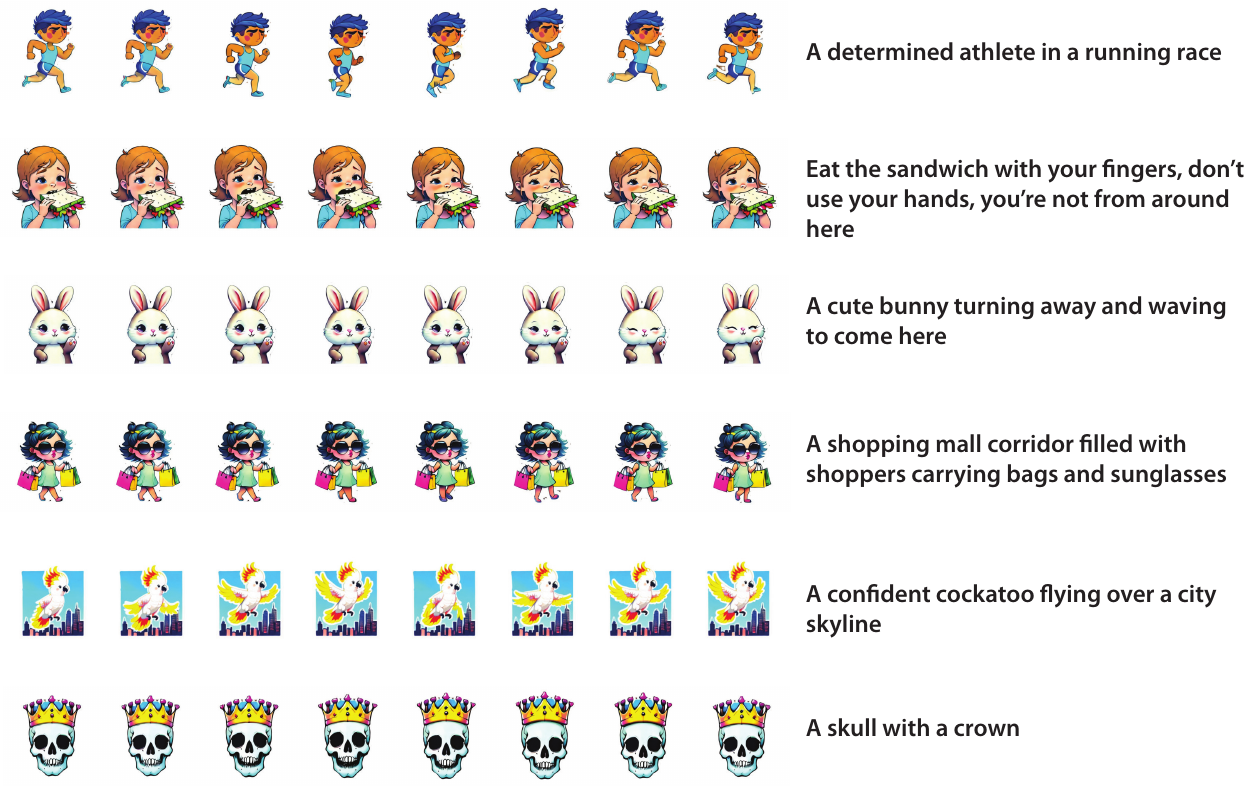}
    \caption{More examples of animated stickers generated by the optimized, HITL-finetuned model. Examples were selected which had motion which was more obvious when viewed as a series of images. Images are shown with transparent regions replaced by a white background.}
    \label{fig:more_examples}
\end{figure}

Figure \ref{fig:more_examples} shows some additional examples of animated stickers generated by the optimized student model. Since motion is more apparent when viewed as a looped video, examples were selected which had motion that was more visible when viewed as a sequence of images.

\end{document}

%% file: 1.introduction.tex
\section{Introduction}
There has recently been an surge of interest in generative text (and image) to video (T2V) modeling~\cite{blattman2023videoldm, blattmann2023stable, girdhar2023emuvideo, ho2022imagen, rakhimov2020latent, hong2022cogvideo, yu2023magvit, guo2023animatediff}. Generated videos from current state-of-the-art models are typically short (under 3 seconds), and typically use text (text-to-video, or T2V), an image (image-to-video, or I2V), or both. 
In this work, we use a text-and-image-to-video architecture to target a natural application of short video generation: animating stickers for social expression. Our model uses the same kind of text-to-image-to-video generation pipeline as Emu video~\cite{girdhar2023emuvideo}, where the image used as conditioning is produced by a text-to-image (T2I) sticker model~\cite{sinha2023stickers}, which we refer to as the static stickers model. Some examples of our model's outputs can be seen in Figure \ref{fig:teaser}. 

\begin{figure}[h]
  \centering
  \includegraphics[width=1.0\linewidth]{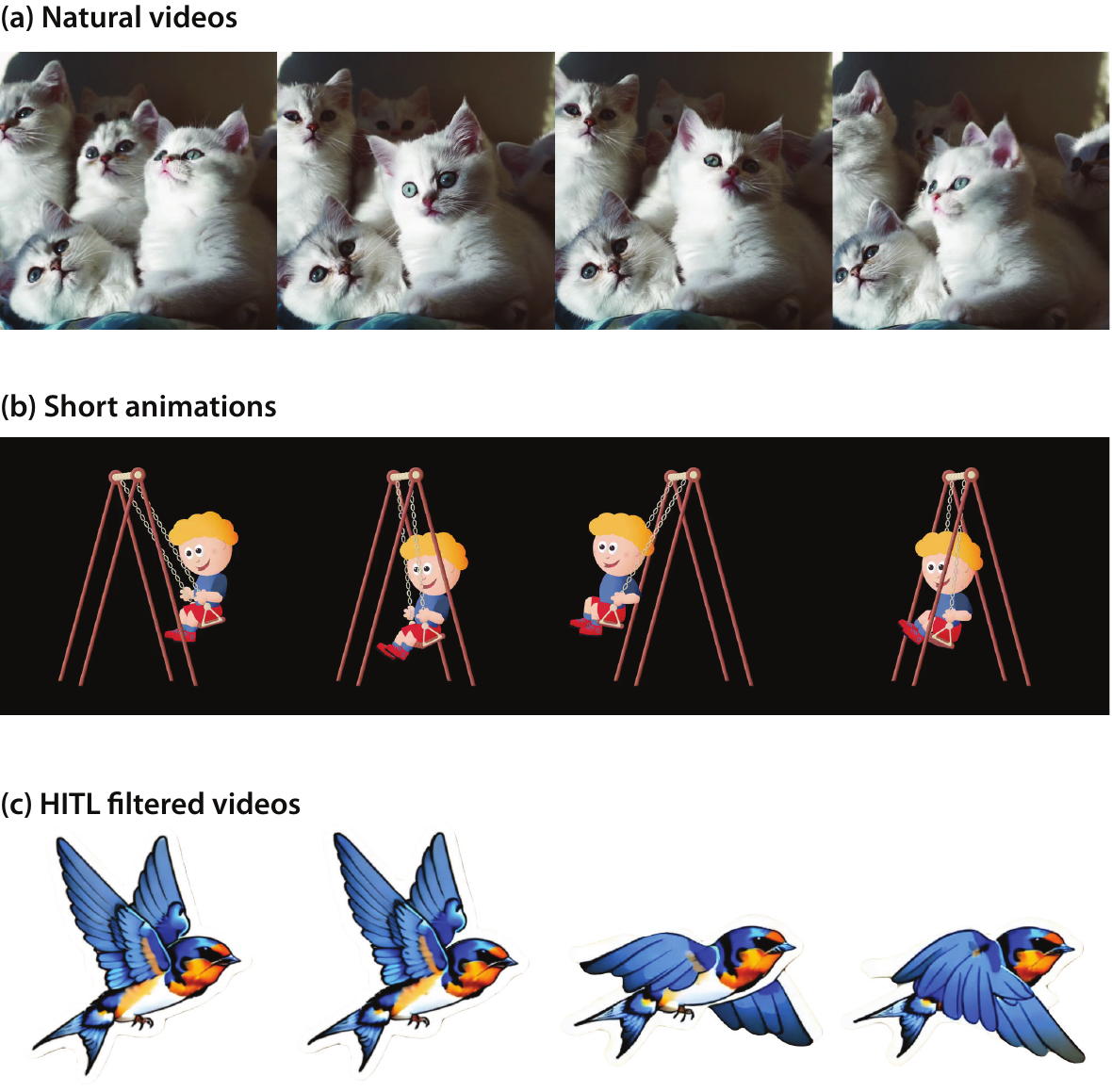}
  \caption{An example of the types of training data used, showing the domain gap between natural videos (a), short animations (b), and HITL-filtered in-domain videos (c).}
  \label{fig:domain_gap}
\end{figure}

Leveraging the existing T2I stickers model achieves the desired style ``for free''. 
However, we find that using a general-purpose I2V model (i.e. one trained only on a general video dataset) does not produce high-quality motion when applied to stickers, and frequently generates videos with static or trivial motion (e.g. only a ``bobbing" effect) and/or introduces inconsistencies and motion artifacts (e.g. warping). 
This is due to the visual and motion differences between natural (photorealistic) videos and sticker-style animations, i.e. a \textit{domain gap}. One example of this gap is that our stickers are entirely contained in the center of the canvas, with a solid color background, which must then be masked to be transparent.
Figure \ref{fig:domain_gap} shows visual examples of our pretrain (natural videos), weakly in-domain data (short animations) and in-domain (human-filtered videos) sets, highlighting the large differences between the source (natural videos) and target (sticker animations) domains.

In this work, we bridge the domain gap using an ensemble-of-teachers human-in-the-loop (HITL) training strategy. 
First, a number of ``teacher'' models are trained using different ``recipes'' of datasets and frame sampling rates, so that collectively, the teacher models are capable of producing high quality diverse motion, though only rarely. 
Next, an HITL dataset is constructed by performing inference using teacher models on a large prompt set covering a wide range of concepts and motions, and then filtered manually for videos with the highest quality motion. 
``Student'' models are then trained directly using the HITL dataset. 
This two-stage approach produces better results than single-stage finetuning on short animations, regardless of quality of the dataset used for single-stage finetuning.

Our model is intended for use in production, and so needs to be fast at inference-time, without sacrificing visual or motion quality. To speed up the model, we utilize three approaches: first, we allow student models to be architecturally smaller than teacher models, using fewer weights and/or fewer text encoders. Second, we use a variety of optimizations which don't require training, i.e. lowering floating point precision, reducing the number of model evaluations needed during sampling, and serializing the model with Torchscript. Finally, we use model distillation to even further reduce the number of sampling steps and model evaluations. The optimized model produces eight frames of four-channel video (RGB plus transparency) in less than 1 second per batch on an H100 GPU, with minimal degradation in quality, and is currently deployed for public testing.

In summary, our contributions are as follows:
\begin{enumerate}
    \item We present our end-to-end process for creating, training, finetuning and optimizing a domain-specific generative video model
    \item We describe our ensemble-of-teachers HITL finetuning strategy, and show that it dramatically improves motion quality and relevance
    \item We describe two video-specific train-time improvements to the data and model -- middle frame conditioning and motion bucketing, and show that these improvements further increase model quality
\end{enumerate}

\subsection{Related work}
\textbf{Video generation models.} With the success of diffusion-based image generation models, there has been a great deal of recent work in generating videos. Video generative models take conditioning signals from text \cite{singer2022make, girdhar2023emuvideo, blattman2023videoldm, ho2022imagen, khachatryan2023text2videozero, hong2022cogvideo, yu2023magvit}, images \cite{girdhar2023emuvideo, guo2023animatediff, yu2023magvit}, sketches \cite{dhesikan2023sketching}, pre-extracted depth maps and optical flows \cite{liang2023movideo}, and others, and generate videos which align with their respective conditioning. In general, video generation models can be classified into those based on generative adversarial networks (GANs) \cite{clark2019adversarial, luc2021transformationbased, aldausari2020video}, auto-regressive models \cite{villegas2022phenaki}, transformer-based models \cite{yu2023magvit, rakhimov2020latent, hong2022cogvideo} and diffusion-based models \cite{singer2022make, girdhar2023emuvideo, blattman2023videoldm}. In this work, we use a diffusion architecture due to its wide applicability in text-to-image (T2I) \cite{rombach2022highresolution}, text-to-video (T2V) \cite{singer2022make,girdhar2023emuvideo,blattman2023videoldm}, video editing \cite{zhang2023avid,wang2023videocomposer}, text-to-3D \cite{poole2022dreamfusion} and text-to-4D \cite{singer2023textto4d,ling2024align}, as well as its capability of generating diverse outputs with high fidelity.

Diffusion models generate images and videos by adding noise to an input and learn to iteratively denoise using neural network predictions \cite{dhariwal2021diffusion, ho2020denoising, nichol2021improved, shaul2023bespoke}. Latent diffusion models (LDMs) operate in latent space, reducing the heavy computational burden from training on a high-resolution pixel space. In this work, we train a latent video diffusion model \cite{blattman2023videoldm} on our video datasets. We follow the factorized design from Emu-video \cite{girdhar2023emuvideo}, where video generation is decomposed into static image generation given the text prompt followed by generating a video conditioned on the image and prompt.

\textbf{Finetuning to bridge domain gaps.} The training dataset plays a key role in determining the quality of generative AI models. Emu Video \cite{girdhar2023emuvideo} shows that the motion of the generated videos can be improved by finetuning the model on a small subset of high motion and high quality videos. However, as mentioned previously, the key challenge we are facing is the domain gap between real videos and animated stickers. Furthermore, AnimateDiff \cite{guo2023animatediff} points out that collecting sufficient personalized videos on target domains is costly; they instead train a generalizable motion module on a large video dataset and plug it into the personalized T2I to stay faithful to a specific domain. While the paper aims to generate valid animations in different personalized domains, they observed failure cases with apparent artifacts and inability to produce proper motion when the domain of the personalized T2I model is too far from that of the real video dataset. Animated stickers is one such case, as the image domain we are trying to animate is from a previous static stickers model \cite{sinha2023stickers}, which are specifically personalized with dedicated style controlling.

Although training on general animations is a way to learn motion priors on animations, it is still not sufficient to close the domain gap with our desired sticker style. DreamBooth \cite{ruiz2022dreambooth} attempts to close the domain gap by using a rare string as the indicator to represent the target domain and augments the dataset by adding images generated by the original T2I model. Here we follow the same inspiration by taking advantage of high quality generated videos in the target domain. We first finetune several teacher models on animation videos to try to learn motion prior on animation style. Then we apply those teacher models to generate videos conditioned on sticker image. Instead of blending those generated videos with animation finetuning videos, we only use generated video clips in sticker style to finetune a student model. This one-hop domain transfer from pretrained model proves to stabilize motion prior in sticker space, without catastrophic forgetting of the motion prior learned from the original pretraining set.

%% file: 2.model_data.tex
\section{Model and Data}
\subsection{Model architecture}
\begin{figure*}[ht]
    \centering
    \includegraphics[width=0.85\linewidth]{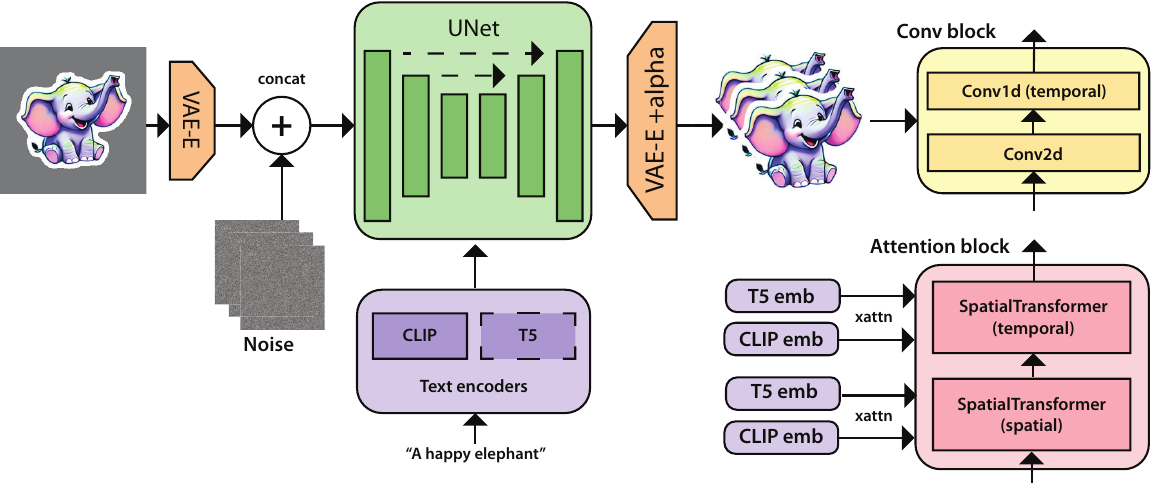}
    \caption{Overall architecture of our animated stickers model (left), and addition of temporal layers to transformer and convolutional blocks (right). We employ a spatiotemporal latent diffusion model (LDM), where  The UNet consists of convolutional stages and attention stages, where the attention stages perform both self and cross-attention to text embeddings (CLIP is always used, FLAN-T5XL is optional depending on the architecture). Temporal layers are added after convolution and spatial transformers, with identity-initialization so that a newly initialized model can load T2I weights and reproduce the T2I model. 
    }
    \label{fig:model_arch}
\end{figure*}

Figure \ref{fig:model_arch} shows an overview of our model architecture. We employ a latent diffusion model (LDM) with additional temporal layers, similar to the approach taken by VideoLDM~\cite{blattman2023videoldm}. Practically, our model architecture is the same as Emu-Video~\cite{girdhar2023emuvideo}, which itself extends Emu~\cite{dai2023emu}, with the only difference being that we use an IP2P-style conditioning instead of masked conditioning. We briefly summarize our model architecture here.

Our model consists of a variational autoencoder (VAE), UNet, and one or more text encoders. The UNet uses the layers and weights from Emu~\cite{dai2023emu}, with 1D convolutions across the time axis inserted after each 2D convolutional layer in ResNet blocks, and temporal attention layers inserted after each spatial attention block. Temporal layers are identity-initialized, so that a newly initialized model with only T2I weights can exactly reproduce text-to-image generation. We use the same VAE as the text-to-image model, including the alpha-channel prediction in \cite{sinha2023stickers}, which allows animations with transparent backgrounds. Our model uses two conditioning signals. Image conditioning is applied by cloning the image latent across the time dimension and appended along the channel axis to the noise, in a similar way as InstructPix2Pix (IP2P)~\cite{brooks2022instructpix2pix}. Text conditioning is applied by encoding a prompt using CLIP~\cite{radford2021learning} and Flan-T5-XL~\cite{chung2022scaling} (where the latter may be omitted in more efficient architectures), and fed into cross-attention layers. As we use two conditioning signals, we enable classifier-free guidance (CFG) by dropping text and image conditioning each separately between 5 and 10\% of the time and together between 5 and 10\% of the time during training, and use the IP2P CFG formulation at inference,
\begin{align}
\tilde\epsilon_{\theta} (z_t, c_I, c_T) = &\epsilon_{\theta} (z_t, \varnothing, \varnothing) \\
&+ \sigma_I (\epsilon_{\theta} (z_t, c_I, \varnothing)) -\epsilon_{\theta} (z_t, \varnothing, \varnothing)) \\
&+ \sigma_T (\epsilon_{\theta} (z_t, c_I, c_T)) -\epsilon_{\theta} (z_t, c_I, \varnothing))
\end{align}
where $z_t$ is the noisy latent, $c_I$ and $c_T$ are the image and text conditionings, respectively, and $\sigma_I$ and $\sigma_T$ are the image and text classifier-free guidance scales. In practice, we use $\sigma_I$ in the range 7.5 to 9 and $\sigma_T$ in the range 1.5 to 3. 
\subsection{Pretraining data} \label{pretraining_data}

35 million natural videos from Shutterstock were used for pretraining video models. Data used for in-domain finetuning included two large (15-60k) short animation datasets, as well as a high quality, professionally-animated sticker set.
\begin{itemize}
\item \textbf{Keyword Based Sourcing + Manual Filtering}. We initially used keyword matching to source 15000 animated videos. These were then manually downselected for samples which were on-style and had high motion quality, resulting in a dataset of 4000 sticker-style short videos. 

\item \textbf{Artist set}. We collected a set of artist-animated sticker packs, and manually removed stickers with text overlaid. In total, 1829 animated stickers were curated. Though the artist dataset has the highest quality in-domain videos (created by professional creatives specifically to be used as social stickers), even this set contains some examples with low quality motion, such as very fast ``jittering'' or videos which alternate between only two frames. These types of motion make sense in the context of their respective sticker packs, but were undesirable for our model, so we further manually filtered out 20\% of videos from this set.

\item \textbf{Video KNN}. To further expand pretraining data, we used video embeddings to perform KNN searches of short videos, using the human-curated sticker-style videos and artist set as seeds. This resulted in a further 62000 medium-to-good quality animated sticker videos. We used an internal video understanding model that was trained using temporal attention and considered different modalities like visual frames, audio, OCR and other signals to produce a multimodal video embedding. We experimentally validated that this model significantly outperformed other simpler video embeddings extracted using only the thumbnail of the video or just visual frames.
\end{itemize}

The artist set has human-written captions which provide detailed descriptions of both motion and content, but the original captions for KNN and keyword-sourced videos tend to be far noisier, and often do not describe the video. To improve these captions, we utilize an internal video captioning model and an entity extraction model. Specifically, we train a video captioning model bootstrapped from the BLIP model~\cite{li2022blip} (trained with the Shutterstock image dataset and additional data) using the divided spatial-temporal self-attention mechanism from~\cite{bertasius2021space}. We adopt a two-stage training strategy to train our video captioning model: a pre-training stage using the Shutterstock dataset, and a finetune stage using the animated videos from Shutterstock and the aforementioned artist dataset. To extract the named entities from each video's original caption, we leverage an entity linking system built on the knowledge base of Wikipedia. By concatenating the outputs from both models, we are able to generate richer descriptions that capture both motion and visual objects in detail.

\subsection{HITL data}
Data for human-in-the-loop (HITL) was created by first curating a set of 15000 prompts, and then sending the prompts into the static stickers model to generate two images per prompt. The prompts came from two main sources: a previous static sticker HITL prompt set, and generations using LLAMA \cite{touvron2023llama}. The prompts generated from LLAMA were curated to describe dynamic motions in order to optimize for large motion in the animated stickers. 

The prompt-image pairs were then used to generate videos using a number of teacher models, the details of which will be discussed in Section \ref{training_details}. Generated videos were sent to human annotators for filtering that fulfilled shareability guidelines, which are defined by three main criteria: 
\begin{itemize}
\item \textbf{Motion quality}. Motion quality is defined as the amount of motion, smoothness of the motion, and if the motion is natural and expressive. A shareable animated sticker will have large motions that is smooth and natural.
\item \textbf{Relevance}. Relevance looks to see if the purpose of the animated sticker is clear with no room for misinterpretation. The movement in the animated sticker is expected to be related the subject and prompt. 
\item \textbf{Consistency}. A shareable animated sticker should not distort or morph in any way. 
\end{itemize}

\begin{figure}[h]
    \centering
    \includegraphics[width=1.0\linewidth]{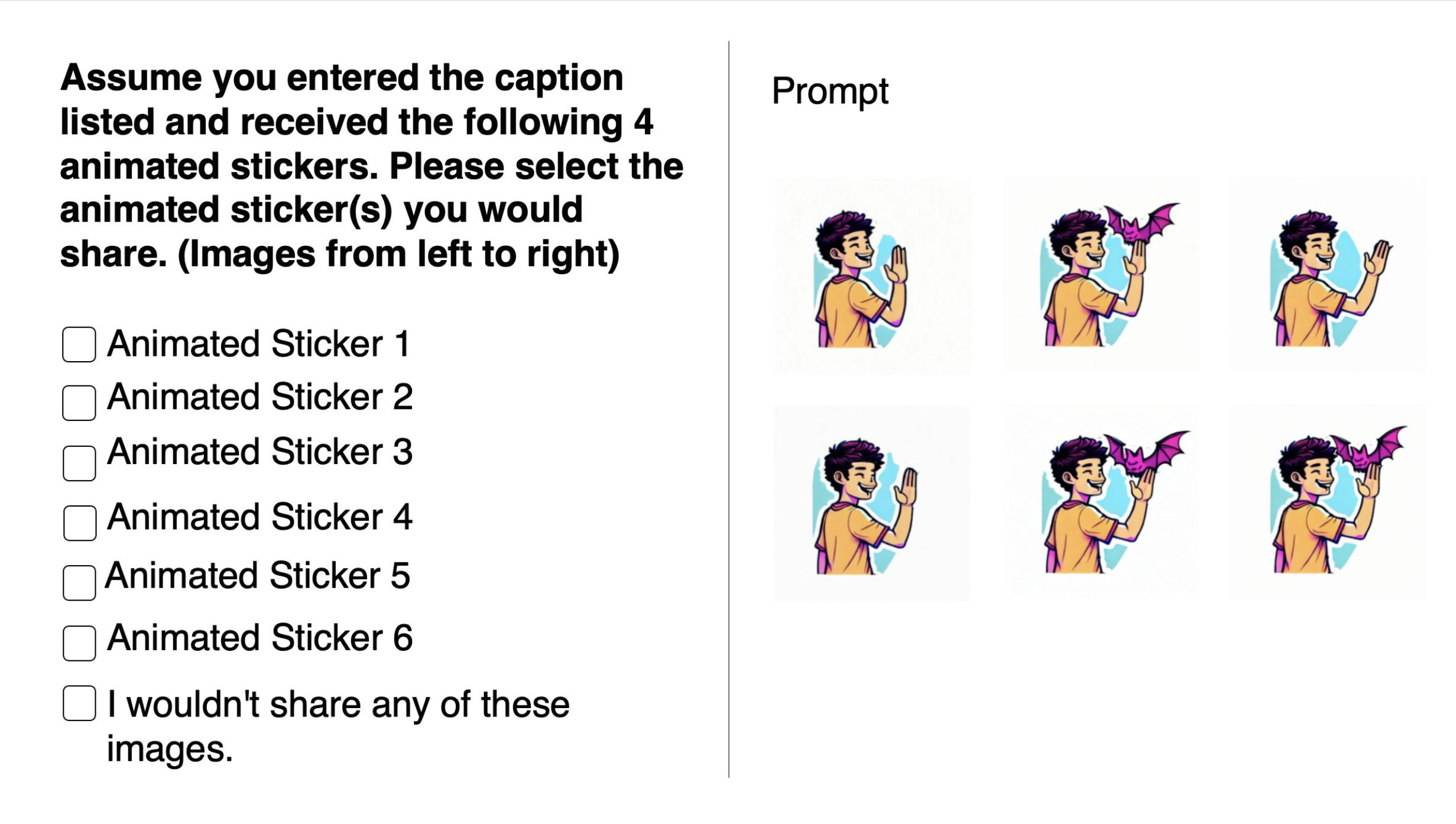}
    \caption{A mock-up of the annotation interface. To the left, annotators select any number out of the available videos, or select ``I wouldn't share any of these image'' if none of the videos are acceptable. To the right, annotators can see the caption, and auto-looped animated sticker videos.}
    \label{fig:hitl}
\end{figure}

A first round of filtering was performed by trained third party vendors with two-out-of-three annotator agreement, followed by a second round of filtering done by internal experts, to ensure the highest quality data. Each round of filtering included jobs that showed six different animated stickers. The annotators were instructed to select all of the animated stickers that fulfilled the shareability criteria. Figure \ref{fig:hitl} shows an example of the interface that the annotators saw. The final selection of animated stickers was used to train the student models.

The prompts were split into three different buckets: \textit{action prompts}, which focused on various actions, ranging from simple actions, such as "a person waving", to dynamic actions, such as "a speeding car navigating a winding road", \textit{emotion prompts}, which capture a wide variety of emotions, and ranged from simple prompts, such as "a girl crying", to complex prompts, such as "a dejected-looking puppy with its ears drooping and its tail between its legs", and \textit{open-ended prompts}, which describe any prompts that do not fall into the emotion and action prompt buckets, such as those about scenery and single=word prompts. In total, out of 15000 starting prompts, about 1500 remained in the post-human-filtering HITL training set.

%% file: 3.training_details.tex
\section{Training details} \label{training_details}
\subsection{Pretraining} \label{pretraining}

After initializing with text-to-image weights, we pretrain our I2V architecture using a 35M Shutterstock video dataset. We find that the highest quality general-purpose I2V models are ones which are trained using a multi-stage process, where at each stage, we change one or more of the following hyperparameters, in addition to tuning normal training parameters such as learning rate and number of training iterations: whether spatial weights are frozen or unfrozen, the spatial resolution (256p or 512p), frame sample rate (either 4 or 8 fps, or dynamic -- see Section \ref{motion_bucketing}), and which quantity the UNet predicts, either the noise $\epsilon$ or the phase velocity $v$ \cite{salimans2022progressive}. Additionally, when using $v$ prediction, we always rescale the noise schedule for zero terminal SNR. \cite{lin2024common}. An example training recipe is [256p, freeze spatial, 4 fps, $\epsilon$-prediction] $\rightarrow$ [512p, freeze spatial, 4 fps, $\epsilon$-prediction] $\rightarrow$ [512p, unfreeze spatial, 8 fps, $v$-prediction].

Using different training recipes such as this one allows us to trade off between motion size and consistency. Empirically, we find that training with $\epsilon$-prediction in early stages increases motion size, and starting from a smaller spatial resolution increases motion quality of the final model. We always train with $v$-prediction and zero terminal SNR in the final stage, as videos generated with $v$-prediction have dramatically better color saturation compared to $\epsilon$. We trained our models on A100 and H100 GPUs with batch size between 128 and 512, learning rate between $2.5\text{e-}5$ and $1\text{e-}4$, and number of iterations between a few thousand and a 150 thousand, depending on whether we were finetuning or pretraining. Videos were resized and center-cropped during training, and we randomly selected 1-second (sample rate of 8fps) or 2-second (sample rate of 4fps) clips and uniformly sampled 8 frames from the clips as training examples.
\subsubsection{Motion bucketing} \label{motion_bucketing}
When sampling training clips from videos, we usually sample all videos from a dataset at the same framerate, with uniform spacing between frames in the clip. For example, when sampling a 24 FPS video at 4 frames per second, we sample every sixth frame, with the general spacing between frames given by $\text{min}\left(\text{round}\left(\frac{\text{video fps}}{\text{desired fps}}\right), \left\lfloor\frac{\text{video frames}}{\text{desired frames}}\right\rfloor\right)$.

However, real-world video datasets will typically contain videos with artificial speed-ups and slow-downs. Additionally, the true level of motion varies widely between videos, and even between different parts of the same video. For applications like sticker animation, a consistent level of motion (neither too much or too little) is key, so we introduced a method to normalize sampling frame rate against actual motion size.

To do this, we compute a motion score for a video, then put scores into FPS ``buckets'' via manual inspection of videos within each bucket. For a first version of this score, we used the vmafmotion \cite{li2018vmaf} score, which is a measure of the temporal difference between adjacent frames; for an updated version, we averaged the norm of the motion vectors from H.264/MPEG-4 AVC standard \cite{KWON2006186}, which are designed for inter-prediction of macroblock offsets to reference frames, over all frames. FPS bucketing results in a mapping between scores and sampling FPS, which we use to dynamically sample videos at train-time. This method is only applicable to longer videos, where it is possible to sample at different framerates -- the HITL data, for example, has only eight frames and does not permit motion bucketing.

Practically, we find that in-domain fine-tuning with motion bucketing improves motion consistency and reduces variance in motion size.
\subsubsection{First vs. middle frame conditioning}
When choosing which frame to use as conditioning during training, the most obvious choice is the first frame. That is, when sampling clips from videos at train-time, use the first frame of the sampled clip as image conditioning. However, we must also consider that, first, at inference-time, the image generated from a prompt with an action (e.g. two people high-fiving) will typically render an image depicting the middle or end of the action. Second, generated frames further in time from the conditioning frame have been empirically found to be more likely to be inconsistent or introduce artifacts. For these reasons, we experimented with using the middle frame (in practice, the fourth frame out of eight) as image conditioning, and find that motion consistency is improved.

Other possible choices for frame conditioning are last frame, and randomly selecting a frame. When experimenting with these, we found that using the last frame gave similar results as using the first frame, and using a random frame gave noticeably worse results. A visual comparison between first-frame and middle-frame model generations is shown in Section \ref{first_vs_middle}.
\subsection{Ensemble-of-teachers HITL} \label{ensemble_of_teachers}
\begin{figure}[h]
    \centering
    \includegraphics[width=\linewidth]{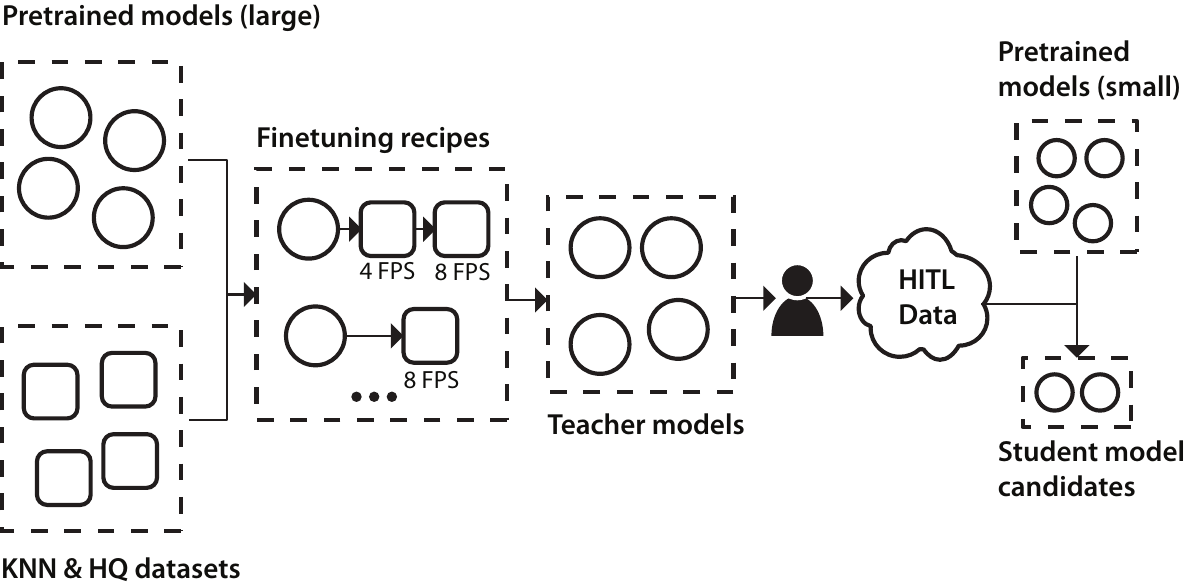}
    \caption{Ensemble-of-teachers finetuning, where a number of pretrained, large general-purpose video models are finetuned using finetuning data and different recipes, which vary by data order and sampling framerate. This results in a set of ``teacher'' models, which are used to generate videos with the HITL prompt set. After human filtering, high-quality HITL data is used to finetune a set of small, efficient pretrained models and downselected into student model candidates.}
    \label{fig:ensemble_of_teachers}
\end{figure}

Static stickers used a human-in-the-loop (HITL) finetuning strategy to improve text faithfulness and style adherence. Since the style and text faithfulness for the content of the video is overwhelmingly determined by the image used as conditioning, we wanted to use an HITL approach tailored specifically to improve motion quality and diversity. Our HITL finetuning strategy has three objectives:
\begin{enumerate}
\item Distill high quality motion from large models into smaller models, for efficient inference
\item Bridge the domain gap between the pretrained models, which were trained on general videos, and static stickers
\item Maximize the diversity, relevance, and interestingness of animated sticker motion
\end{enumerate}
We take an approach we call ensemble-of-teachers HITL finetuning, which we outline in Figure \ref{fig:ensemble_of_teachers}. This approach differs from the HITL used for static stickers in two ways:
\begin{enumerate}
\item We use multiple expert-selected models to generate the HITL data for human annotation
\item The models which generate the HITL data have different architectures (typically larger) than the models which train on it
\end{enumerate}
We begin with several pretrained foundational models, selected for different levels of motion size vs. consistency, and finetune them using finetuning recipes on the datasets discussed in Section \ref{pretraining_data}. This produces a number of teacher models which are then downselected by human experts according to two factors:
\begin{itemize}
\item \textbf{High peak quality}. Since we were human-annotating a large number of generations for each model, we judged models on their best generated videos, rather than the average. This allows us to filter for very high quality, interesting animations from each teacher model even if the majority of the videos produced by that model are poor.
\item \textbf{Diversity of motion}. We want teacher models to have minimal overlap between each other in terms of motion speed, novelty, and quality in different actions. For example, one teacher model may be great at producing running and walking motions but poor at others. Notably, we are able to train on the same data while sampling at different framerates, so that models trained at different framerates have different distributions of motion speed.
\end{itemize}
These teacher models are used to generate videos from an HITL prompt and image set  which is filtered by human annotators, engineers, and creatives (covered in the previous section). The downselected high quality HITL set is then used to train a number of pretrained student models, some of which may be architecturally different than the teacher models -- these architectures are covered in detail in Section \ref{efficient_architectures}. Finetuning on data which is more aligned with the output distribution makes the model generate more stable, consistent, and higher quality motion. Also, teacher models trained on noiser data (e.g. the KNN data) often produces large but low-quality and inconsistent motion. However, this is tolerable for the purposes of HITL, since we can filter for ``lucky'' instances where the motion is both large and consistent.
\subsection{Model optimizations}
Since the animated stickers model is intended to be used in production, it needs to perform inference quickly in addition to producing high quality motion. We applied three strategies to trade off between inference time and quality: training-free optimizations, reducing the number of UNet weights, and model distillation. These are detailed in the remainder of this section.
\subsubsection{Training-free optimizations}
We first employed some general optimizations which are applicable to any latent diffusion model at inference, independent of architecture or distillation. This included: 
\begin{itemize}
    \item \textbf{Halving the floating point precision}. Converting the model from Float32 to Float16 speeds up the inference time for two reasons. First, the memory footprint of the model is halved. Second, 16 floating point operations can be executed faster. For all models, we use BFloat16 (a float16 variant with a smaller mantissa) \cite{kalamkar2019study}  for training and inference.
    \item \textbf{Torchscripting and freezing}. Torchscript \cite{devito2022torchscript} is a serialized format for easy deployment of PyTorch models. Converting a model from pure PyTorch to TorchScript involves automatic optimizations that can increase inference speed, such as fusing multiple operations, constant folding, and techniques to reduce the complexity of the computational graph. Additionally, freezing (referring to \texttt{jit.freeze}, not weight freezing) allows further automatic speed optimizations in Torchscript, by converting dynamic parts of the graph into constants to remove unnecessary operations. Importantly, we freeze with the flag to preserve the numerics and prevent quality degradation. 
    \item \textbf{Optimized temporal attention expansion}. Temporal attention layers (attending between the time axis and text conditioning) require the context tensors to be replicated to match the number of frames (the time dimension). In a naive implementation, this would be done before passing to cross-attention layers. The optimized version takes advantage of the fact that the repeated tensors are identical, and expands after passing through the cross-attention's linear projection layers, reducing compute and memory.
    \item \textbf{DPM-solver}. Rather than use the DDPM \cite{ho2020denoising} or DDIM \cite{song2020denoising} solvers for inference, which typically require more sampling steps for good quality generation, we use DPM-solver \cite{lu2022dpm} and a linear-in-logSNR time schedule at inference to reduce the number of sampling steps to 15.
    \item \textbf{Adaptive guidance}. A novel technique that reduces the number of network evaluations from three to two one for a subset of the sampling steps \cite{castillo2023adaptive}. In effect, less forward passes through the network are executed and memory usage is reduced. These two effects result in faster inference speed without any quality degradation. In practice, we perform full guidance for the first eight (out of 15) sampling steps, and no guidance for the remaining seven.
\end{itemize}

With these optimizations, we are able to reduce inference time by an order of magnitude when compared to a fully unoptimized model (e.g. DDIM 50 steps, full precision, full guidance) with minimal change in quality.
\subsubsection{Efficient model architectures} \label{efficient_architectures}
Our pretrained video model, with CLIP and T5 text encoders, which, as mentioned previously, is roughly the same architecture as Emu Video. To reduce the number of weights, we targeted the following areas:
\begin{enumerate}
\item Number of UNet channels
\item UNet spatial and temporal transformer depth
\item Number of UNet resnet blocks per UNet block
\item Whether to include the T5 text encoder or only use CLIP
\end{enumerate}
Notably, we do not reduce the number of latent channels (which stays at 8 for all experiments), as we empirically found that having at least 8 channels is important to reducing visual artifacts and morphing. As an example, at 512p, the foundational UNet has 4.3 B weights and uses 23.5 teraFLOPs, whereas a more efficient UNet ("sm") has 1.2 B weights and uses 5.6 teraFLOPs.

For our students, we narrowed potential models to four UNet architectures: ``lg'' (4.3 B UNet weights), ``lg-e'' a.k.a. lg-efficient (fewer res blocks, and no T5 encoder, 3.5 B UNet weights), ``med'' (2.4 B UNet weights) and ``sm'' (1.2 B UNet weights). These models were pretrained using similar recipes as used for the teacher models in Section \ref{pretraining}, with the notable difference being student models are all trained up to a maximum of 256p, since that is the required output size.
\subsubsection{Distillation}
To speed inference up further, we use two distillation techniques that reduce the number of forward passes through the UNet without affecting the parameter count: 
\begin{itemize}
\item \textbf{Guidance distillation}. 
Diffusion models use classifier-free guidance for conditional image generation, which requires a conditional and unconditional forward pass per solver step. Guidance distillation reduces these two forward passes into one. However, in the case of the animated stickers model, classifier-free guidance requires three forward passes per step: a full conditional (text and image), unconditional, and an image-conditional step. Applying guidance distillation to reduce three forward passes into one has not yet been described in the literature, but we found that it works well in practice, reducing inference time threefold.
\item \textbf{Step-distillation}. 
In step distillation, a teacher and student are initialized with the same weights, and the student is trained to match multiple teacher steps in a single step.
\item \textbf{Guidance+step distillation}. We combine guidance and step distillation can by training a student to imitate classifier-free-guidance and multiple steps at the same time with just one forward pass through the UNet. We find that a four-to-one ratio of teacher to student steps works best. Distilling 32 teacher steps into 8 student steps during training. Our final model only requires eight solver steps, with one forward pass through the UNet per step.
\end{itemize}

For each of the four efficient UNet architectures (sm, med, lg, lg-e), we evaluated training-free optimization, guidance distillation, and guidance+step distillation. Benchmark times for the models ranged from 300 ms on an H100 for the smallest model with both guidance and step distillation, to 2000 ms for the largest model with only training-free optimizations.

Ultimately, we select the lg model with guidance+step distillation, which has an H100 inference time of 726 ms, for evaluation and public testing as the best compromise between inference time and quality. Heavily-distilled smaller models were found to have more frequent artifacts and worse motion, and more expensive models had slightly better motion, but at a too-heavy computational cost. The models with only training-free optimizations were most faithful to the original model, but still significantly slower than the distilled models.  

%% file: 4.evaluation_results.tex
\section{Evaluation and results}
\subsection{Evaluation}
In order to evaluate the quality of the model, we created an annotation guideline to preform standalone evaluations for the different versions of the animated sticker model. Standalone evaluations mean that we show the annotators one animated sticker. The annotation guideline provides questions to evaluate animated stickers based on motion quality and prompt similarity. Annotators were instructed to place a stronger focus on motion quality due to the fact that prompt similarity strongly correlates with the content of the sticker image used for conditioning. 

The motion quality task has ten questions pertaining to the motion in the animated sticker: 
\begin{enumerate}
\item Existence: Is there any motion in the animated sticker?
\item Relevance: Is the motion in the animated sticker expected and relevant to the prompt?
\item Consistency: Do the characters and objects remain in proportion and consistent throughout the animation?
\item Distortions: Is the animated sticker free of any flicker or distortion?
\item Motion curves: Does the animation utilize smooth motion curves that reference real physics/ gravity?
\item Outline: Does the linework/ white outline of the sticker move with the motion of the sticker? 
\item Looping: Does the animation play in a continuous loop?
\item Clipping: Are there no integral elements of the animated sticker clipped?
\item Expression: Are the expressions in the animated sticker clear?
\item Background: Does background motion complement and not distract from the primary animation?
\end{enumerate}
For each question, the annotators were instructed to either select "yes" or a reason why the animated sticker failed; they were able to select multiple failure reasons.

The prompt similarity task has four questions pertaining to how well the animated sticker matches the prompt. 
\begin{enumerate}
\item Subjects: Does the animated sticker clearly represent all subjects or objects intended in the prompt?
\item Actions: Does the animated sticker effectively depict all specific actions or movements, as stated in the prompt? 
\item Composition: Does the animated sticker depict the desired arrangement and quantity of the subjects or objects specified in the prompt?
\item Attributes: Does the animated sticker include all attributes (e.g., colors, shapes, sizes) and emotions of subjects or objects specified in the prompt?
\end{enumerate}
Similar to the motion quality task, the annotators were instructed to either select "yes" or a reason why the animated sticker failed. The annotators were also instructed to fail the animated sticker if one of the frames failed for the question.

Table \ref{tbl:evals} show standalone evaluation results with three annotator multi-review for the optimized student model. Some of the criteria, such as existence and relevance, have high pass rates because the guidelines for these criteria were not strict. For example, the raters were asked to choose "yes" for the existence of motion as long as there was primary motion in the sticker, which is defined as the main movement in the animation. The animated sticker did not need to have secondary motion, which are animation that amplifies the action by supporting the primary motion with secondary characteristic detail, in order to pass. However, if we tighten the guidelines and require both primary and secondary motion, the pass rate will fall to 0.857. There are plans to tighten the guidelines for future iterations.

We also observe that distortions and consistency have the lowest pass rate out of all of the criteria. We hypothesize that more motion have a larger chance of having distortions and less consistency, which lead to a lower pass rate.

\begin{table}[]
\begin{tabular}{lll}
\hline
Category      & Consensus count & Pass rate \\ \hline
Existence     & 1890                       & 0.969   \\ 
Relevance     & 1928                       & 0.992   \\ 
Consistency   & 1772                       & 0.786  \\ 
Distortions   & 1800                       & 0.673  \\ 
Motion curves & 1888                       & 0.934  \\ 
Outline       & 1894                       & 0.920  \\ 
Looping       & 1894                       & 0.999  \\ 
Clipping      & 1894                       & 0.994  \\ 
Expression    & 1894                       & 0.954  \\ 
Background    & 1928                       & 0.999  \\ \hline
\end{tabular}
\caption{Standalone evaluation results on the optimized student model for all evaluation categories. Consensus count refers to samples where all three annotators agreed on the label, and pass rate is the percentage of samples with consensus where the animated stickers passed the criteria.} \label{tbl:evals}
\end{table}

\subsubsection{Effect of Distillation on Model Quality}
In order to demonstrate the effects of distillation on the model quality, we ran evaluation on the distilled and non-distilled (normal Pytorch model with DDIM 50 steps sampling) versions of the final student model. Table \ref{tbl:evals_distillation} show evaluation results for both versions. While existence, motion curves, and clipping has a slightly higher pass rater for the non-distilled student model, the distilled student model has a higher pass rate for all the other categories. Furthermore, it has much higher pass rate for consistency, distortions, outline, and expression.

\begin{table}[]
\begin{tabular}{lll}
\hline
Category      & Non-distilled Pass Rate & Distilled Pass Rate \\ \hline
Existence     & 0.978                   & 0.969               \\ 
Relevance     & 0.978                   & 0.992               \\ 
Consistency   & 0.572                   & 0.786               \\ 
Distortions   & 0.488                   & 0.673               \\ 
Motion curves & 0.977                   & 0.934               \\ 
Outline       & 0.791                   & 0.920               \\ 
Looping       & 0.993                   & 0.999               \\ 
Clipping      & 0.998                   & 0.994               \\ 
Expression    & 0.707                   & 0.954               \\ 
Background    & 0.995                   & 0.999               \\ \hline
\end{tabular}
\caption{Standalone evaluation results for the non-distilled student model and distilled student model} \label{tbl:evals_distillation}
\end{table}

\subsection{Results and Visualizations}
More general examples can also be found in Appendix \ref{more_animation_examples}.
\subsubsection{Pretrained vs. finetuned model}
\begin{figure}[h]
    \centering
    \includegraphics[width=0.8\linewidth]{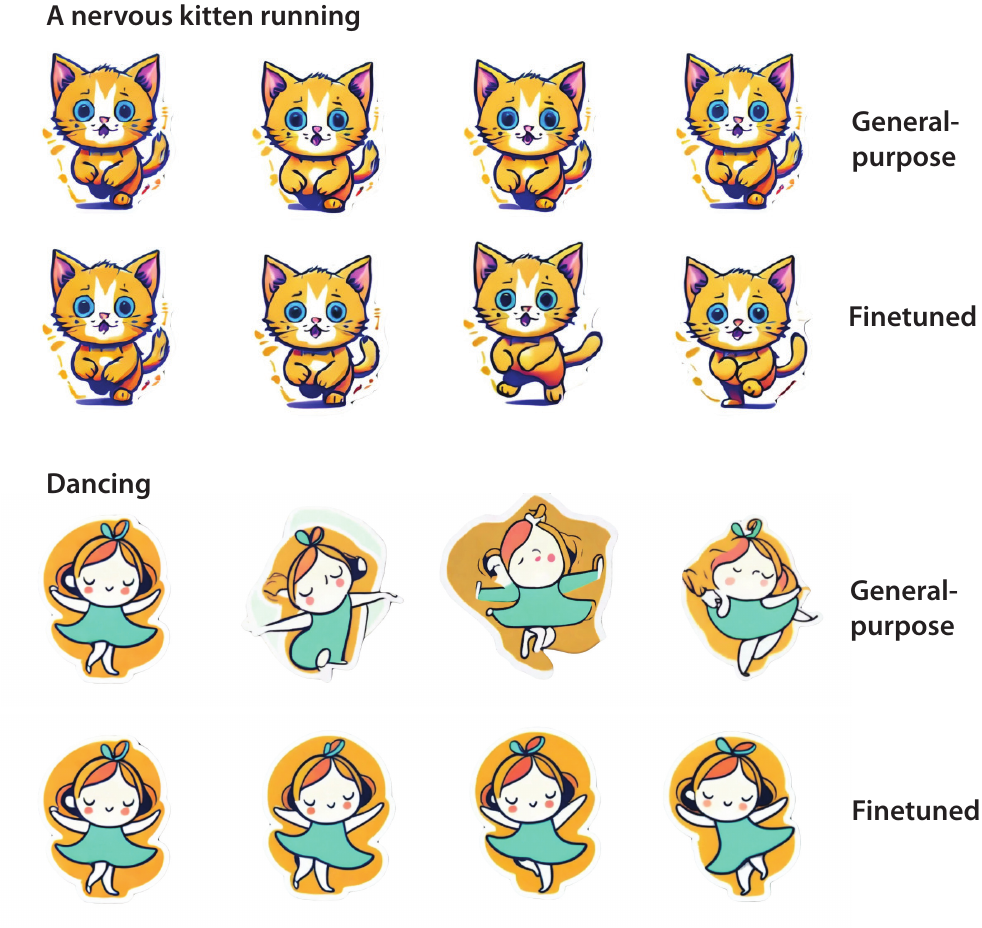}
    \caption{Examples showing the effect of finetuning versus a general-purpose (out-of-domain) video model trained on natural videos. In-domain and HITL finetuning has the effect of a) increasing secondary motion (e.g. in faces, background objects, etc.), b) giving the subject a relevant animation rather than adding a bulk motion, and c) reducing motion artifacts and morphing. Top: the general-purpose model gives the cat an up-and-down bobbing motion, whereas the finetuned model animates a correct running movement. Bottom: the general-purpose model adds morphing to the video, whereas the finetuned model correctly animates dancing.}
    \label{fig:general_purpose_vs_finetuned}
\end{figure}

In order to demonstrate the significant improvement in motion from in-domain and HITL finetuning, we show some examples in Figure \ref{fig:general_purpose_vs_finetuned} of the same image and prompt conditioning, animated with a 256p-trained general-purpose video model (trained on Shutterstock only) versus the student animated stickers model. Note that the general-purpose model is capable of correctly animating natural images.

In addition to adding motion to many stickers where the general-purpose model generates a static video, the full finetuning pipeline makes large improvements in three areas: (1) increasing secondary motion, for example in faces, background objects, and body parts (the general-purpose video model very rarely generates secondary motion in sticker-style videos, which reduces expressiveness), (2) giving the subject a relevant animation rather than bulk motion (e.g. only rotation or translation of the sticker) -- the HITL-finetuned model is able to correctly animate a number of actions, such as running, jumping, laughing, hugging, etc, but even when there isn't full prompt-action alignment, the HITL-finetuned model tends to give the subject correct and relevant motion relative to itself, e.g. limbs and faces moving correctly relative to the body, and (3) reducing motion artifacts, such as morphing and blurring.
\subsubsection{First vs. middle-frame conditioning} \label{first_vs_middle}
\begin{figure}[h]
    \centering
    \includegraphics[width=0.8\linewidth]{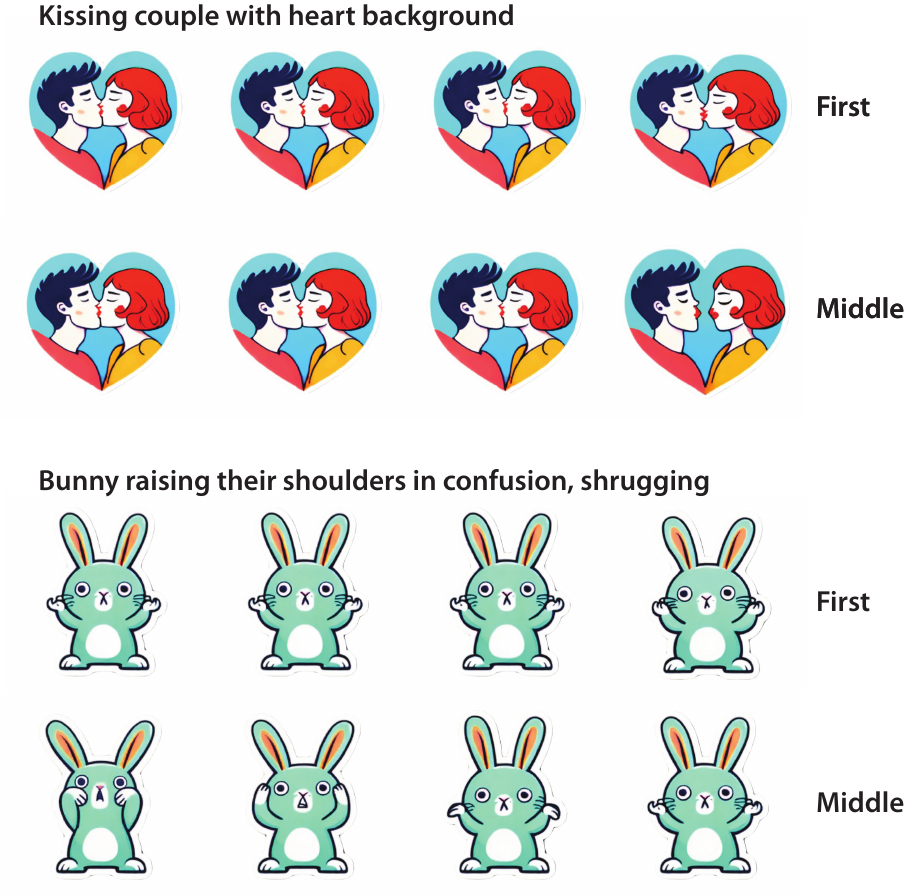}
    \caption{Examples showing the difference between training using the first frame as conditioning and using the middle (fourth) frame as conditioning. In general, as shown in the examples, middle-frame conditioning tends to produce larger motion, since the action in the prompt generally produces a static sticker which depicts the middle of the action rather than the beginning.}
    \label{fig:first_vs_middle}
\end{figure}

To highlight the effects of middle-frame conditioning, we trained two models: one which was pretrained on Shutterstock using and then finetuned on the artist sticker set using first frame conditioning for both pretraining and finetuning, and another with the same pipeline but using middle-frame conditioning for both. Figure \ref{fig:first_vs_middle} shows some comparisons between the two models.
In general, we find that both motion consistency and size are improved with middle-frame conditioning. In the examples in Figure \ref{fig:first_vs_middle}, both cases show larger and more natural motion for middle-frame conditioning, where first-frame conditioning only shows some ``bobbing" motion.